\documentclass[10pt,journal,compsoc]{IEEEtran}

\ifCLASSOPTIONcompsoc
\usepackage[nocompress]{cite}
\else
\usepackage{cite}
\fi

\ifCLASSINFOpdf
\else
\fi

\usepackage{cite}
\usepackage{graphicx}

\usepackage{booktabs}
\usepackage{color}
\usepackage{array}
\usepackage{subfigure}
\usepackage{multirow}

\usepackage{enumitem}

\usepackage{times}
\usepackage{helvet}
\usepackage{courier}
\usepackage[hyphens]{url}
\usepackage{caption}
\usepackage{amsmath}
\usepackage[ruled,linesnumbered]{algorithm2e}

\begin{document}

\title{Meta Auxiliary Learning for Facial Action Unit Detection}

\author{Yong Li,
	Shiguang Shan*,~\IEEEmembership{Member,~IEEE,}% <-this % stops a space
	\IEEEcompsocitemizethanks{\IEEEcompsocthanksitem 
		Corresponding author: Shiguang Shan, sgshan@ict.ac.cn
		\protect\\
		% note need leading \protect in front of \\ to get a newline within \thanks as
		% \\ is fragile and will error, could use \hfil\break instead.
		Yong Li is with the Key Laboratory of Intelligent Perception and Systems for High-Dimensional Information, Ministry of Education, School of Computer Science and Engineering, Nanjing University of Science and Technology, Nanjing, 210094, China (e-mail: yong.li@njust.edu.cn)
		\IEEEcompsocthanksitem S. Shan is with the Key Laboratory of Intelligent Information Processing of Chinese Academy of Sciences, Institute of Computing Technology, CAS, Beijing 100190, China, and with the University of Chinese Academy of Sciences, Beijing 100049, China, and also with CAS Center for Excellence in Brain Science and Intelligence Technology (e-mail: sgshan@ict.ac.cn).}% <-this % stops a space
	\thanks{Manuscript received April 19, 2005; revised August 26, 2015.}}

% The paper headers
\markboth{Journal of \LaTeX\ Class Files,~Vol.~14, No.~8, August~2015}%
{Shell \MakeLowercase{\textit{et al.}}: Bare Advanced Demo of IEEEtran.cls for IEEE Computer Society Journals}

\IEEEtitleabstractindextext{%
\begin{abstract}
Despite the success of deep neural networks on facial action unit (AU) detection, better performance depends on a large number of training images with accurate AU annotations. However, labeling AU is time-consuming, expensive, and error-prone. Considering AU detection and facial expression recognition (FER) are two highly correlated tasks, and facial expression (FE) is relatively easy to annotate, we consider learning AU detection and FER in a multi-task manner. However, the performance of the AU detection task cannot be always enhanced due to the negative transfer in the multi-task scenario. To alleviate this issue, we propose a Meta Auxiliary Learning method (MAL) that automatically selects highly related FE samples by learning adaptative weights for the training FE samples in a meta learning manner. The learned sample weights alleviate the negative transfer from two aspects: 1) balance the loss of each task automatically, and 2) suppress the weights of FE samples that have large uncertainties. Experimental results on several popular AU datasets demonstrate MAL consistently improves the AU detection performance compared with the state-of-the-art multi-task and auxiliary learning methods. MAL automatically estimates adaptive weights for the auxiliary FE samples according to their semantic relevance with the primary AU detection task.
\end{abstract}

% Note that keywords are not normally used for peerreview papers.
}

% make the title area
\maketitle

\IEEEdisplaynontitleabstractindextext

\IEEEpeerreviewmaketitle

\section{Introduction}
\label{sec:introduction}

Facial actions convey various and subtle meanings, including a person's emotion, intention, attitude,  mental and physical states. The most comprehensive approach to encode and annotate facial actions is the  anatomically-based  Facial Action Coding System (FACS) \cite{friesen1978facial}.  FACS defines a unique set of approximately 30 atomic non-overlapping facial muscle actions called Action Units (AUs), which correspond to the muscular activities that produce momentary changes in facial appearance. The combinations of the AUs can describe nearly all possible facial expressions.
Due to their utility for understanding a human being’s mental state, the automatic recognition of facial action units is becoming a popular field in computer vision and holds promise to plentiful applications, such as human-computer interaction, emotion analysis, medical pain estimation.

\begin{figure}[htb]
	\centering
	\includegraphics[width=0.9\linewidth]{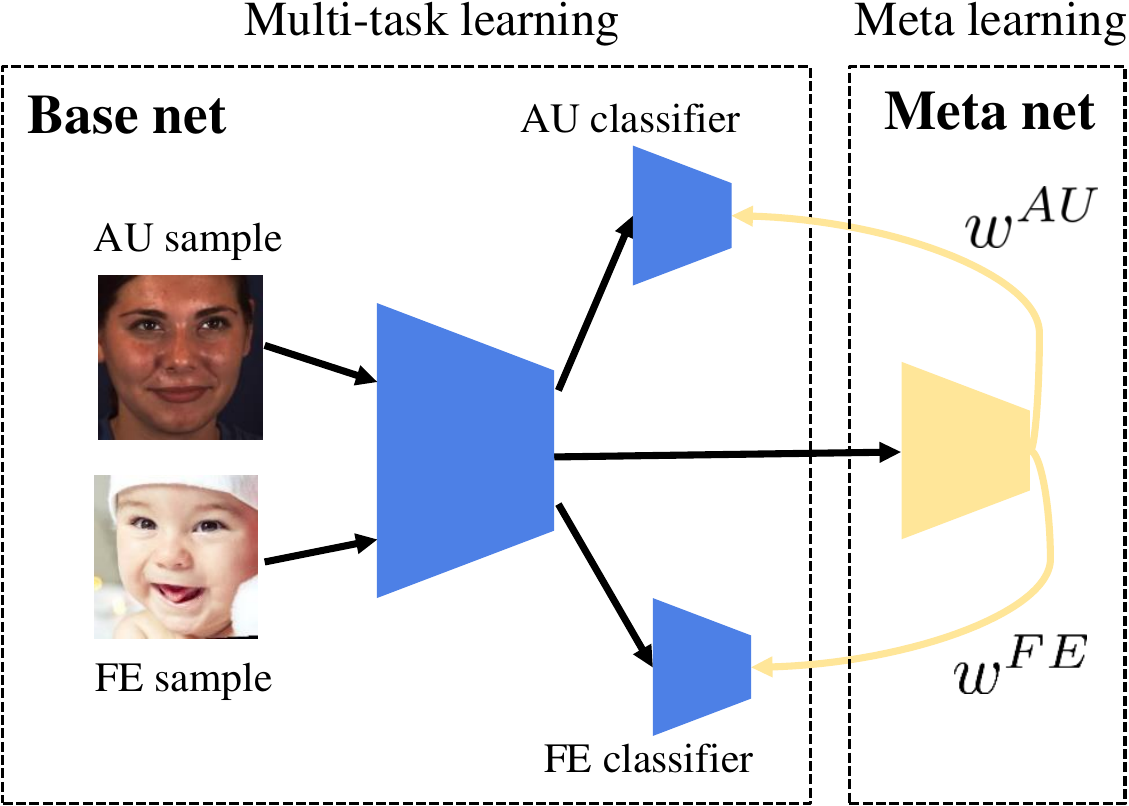}
	\caption{
		Main idea of the proposed Meta Auxiliary Learning method (MAL) for AU detection. 
		The network structure of MAL consists of a base net and meta net that are used for the multi-task learning and sample weight estimation, respectively.
		To adaptatively transfer knowledge from a large amount of FE data for AU detection, MAL leverages the meta net to learn the adaptive sample weights $w^{AU}$ and $w^{FE}$ for AU and FE training samples to minimize the loss on the AU validation dataset.
	}
	\label{fig:main_idea}
\end{figure}

% desc the problem
Recently, deep neural networks (DNNs) have been applied to AU detection and achieved great progress. Such methods are approaching a fundamental limit since existing AU datasets (e.g., BP4D \cite{zhang2013high}, DISFA \cite{mavadati2013disfa}, GFT \cite{girard2017sayette}) merely contains limited subjects as well as facial images. 
Learning DNNs requires a large amount of labeled images, but it is quite time-consuming, expensive, and error-prone to collect a sufficient amount of labeled AU samples.
It is because annotating AU images requires professional expertise, the facial deformation caused by AUs is often subtle and regional, such nuance is further obscured by variations caused by intensities and individuals. As illustrated in \cite{donato1999classifying}, human coders require time-consuming, intensive training and practice before they can reliably assign AU codes. After training, coding photographs or videos frame by frame is a slow process, which makes human FACS coding impractical to use on facial movements as they occur in everyday life. Large inventories of naturalistic photographs and videos—which have been curated only fairly recently would require decades to manually code.

In this paper, we address this challenge of data scarcity by transferring knowledge from a large amount of facial expression (FE) data to benefit AU detection.  Facial expression  and AU  represent two levels of descriptions for the facial behavior \cite{cui2020knowledge}. FER and AU detection are proved to be the semantically correlated tasks in two-fold: (1)  prototypical AUs can be grouped to describe the basic and compound expressions  \cite{fabian2016emotionet, du2014compound, li2013data}. (2)  convolutional neural networks (CNNs) trained for FER are able to model the high-level features that strongly correspond to the facial AUs \cite{khorrami2015deep, liu2013aware, liu2015inspired}. Therefore, previous works \cite{wang2019multi, liu2019end, cui2020knowledge} proposed to learn FER and AU detection in a multi-task manner to enhance the two tasks simultaneously. However, the benefits of the multi-task learning paradigm are not always guaranteed. Individual tasks may experience negative transfer in which the performance of the multi-task model can be worse than that of a  single-task model \cite{wang2019characterizing}. As illustrated in \cite{wang2019characterizing, liu2019loss}, the negative transfer occurs naturally in the multi-task learning scenarios when the source data are less related or one task dominates the training process.

%One one hand, such multi-task methods have to handcrafted tune both the model structure and loss weights. On the other hand, the optimiser in mutli-task learning usually foucus on minimizing on the easier task \cite{li2018learning}. Thus the multi-task learning based model would be sub-optical for AU detection. 

To this end, We propose a Meta Auxiliary Learning method (MAL) to adaptively transfer knowledge from a large amount of FE data to boost the AU detection. To automatically balance the two tasks and filter the less related FE samples, MAL is trained to learn optimal weights for the training samples of the current mini-batch from the two tasks by minimizing the loss on the reserved AU validation dataset.
Fig. \ref{fig:main_idea} illustrates the main idea of MAL. The network structure of MAL consists of a base net and a meta net.
For each training iteration, MAL firstly learns FER and AU detection jointly to update the base net, then learns to learn the optimal sample weights through the meta net by evaluating the performance of the updated base net on the AU validation dataset. 
Different from the conventional offline training-validation paradigm that manually tunes the loss weight of the individual tasks,  MAL conducts validation at every training iteration online to dynamically determine the sample weights of the current mini-batch samples. 
The core idea of MAL is: \textit{the learned sample weights should be capable of consistently minimizing the loss on the AU validation dataset}. 
The benefits of the adaptively learned weights are two folds. Firstly, they automatically balance the AU detection and FER tasks during the training process. Secondly, the weights of the FE samples are estimated according to their semantic relevance with the AU detection task.
% Although facial expression is relatively easy to annotate, it is  difficult to annotate the large-scale FER datasets with high quality due to the uncertainties caused by the subjectiveness of annotators as well as ambiguities of facial expressions.
%In the testing phase, merely the base net is utilized for AUs inference and estimation. 

Our main contributions include: (1) We formulate a novel Meta Auxiliary Learning method (MAL) to adaptively transfer knowledge from a large amount of FE data for the enhancement of the facial AU detection.
(2) We elaborately design a meta optimization algorithm to supervise MAL to learn the reasonable sample weights, which are capable of adaptatively suppressing the negative transfer and balancing the two tasks during the multi-task training process.
(3) Experimental results on several AU datasets show that MAL consistently improves the facial AU detection performance. MAL also outperforms the state-of-the-art multi-task learning approaches and is comparable with current best AU detection methods. Besides, weight visualization results show MAL automatically balances the two tasks during the training process and suppresses the weights of ambiguous FE samples.

\section{Related Work}
\subsection{Facial action unit detection} 
Automatical AU detection has been studied for decades and a number of approaches have been proposed \cite{martinez2017automatic, eleftheriadis2015multi, koelstra2010dynamic, wang2013capturing}.
Previous methods that address AU detection can be classified into four categories: supervised \cite{zhao2016deep, li2017eac, shao2018deep, li2017action}, weakly-supervised \cite{peng2018weakly, zhao2018learning}, semi-supervised \cite{zeng2015confidence, chu2013selective}, and self-supervised \cite{wiles2018self, li2019self} learning  methods. For supervised methods,  DRML \cite{zhao2016deep} adopted a locally connected convolutional layer to learn the facial region-specific AU representation. Methods in \cite{li2017eac, shao2018deep, li2017action} leveraged facial landmarks to extract and learn facial region-specific representation.  Li et al. \cite{li2019semantic} proposed an AU semantic relationship embedded representation learning (SRERL) framework that incorporates Gated Graph Neural Network (GGNN) for AU relationship reasoning and modeling. Han et al. \cite{han2018optimizing} proposed an Optimized Filter Size CNN (OFS-CNN) that can learn specified filter sizes and weights of all convolutional layers for each AU. Zhang et al. \cite{zhang2020region} proposed to encode the location of both positive and negative occurred AUs by directly regressing the centers of the pre-defined heatmaps. However, such supervised approaches suffer from insufficient training data and are usually lack model generalizability.

To alleviate the dependence of laborious AU labeling, the weakly-supervised learning methods usually seek to leverage facial images with incomplete \cite{peng2018weakly} or noisy \cite{zhao2018learning, peng2019dual} AU annotations.
The semi-supervised learning methods incorporate both labeled and unlabeled data by assuming the faces to be aggregated by AUs, or to have a smooth label distribution \cite{wu2017deep}.  The self-supervised learning approaches usually adopt pseudo supervisory signals to learn facial AU representation \cite{li2019self, li2020learning, lu2020self}. Among them, Lu et al. \cite{lu2020self} exploited a triplet-based ranking approach that learns to rank the facial frames based on their temporal distance from an anchor frame, they demonstrated that the encoder learns meaningful representations for AU recognition with no labels.

Considering that FE and facial AUs are semantically correlated, several works \cite{pons2018multi, wang2017expression, liu2013aware, wang2019multi, cui2020knowledge} proposed to learn the two tasks jointly. Among them, cui et al \cite{cui2020knowledge} proposed a constraint optimization method to encode the generic knowledge on expression-AUs probabilistic dependencies into a bayesian network.
Results in \cite{pons2018multi, wang2019multi} suggest that the AU detection and FER tasks can promote each other in the multi-task learning paradigm. However, it is  difficult to annotate the large-scale FER datasets with high quality due to the uncertainties caused by the subjectiveness of annotators as well as ambiguities of facial expressions. 
Some FE images do not show any AUs because the perceived emotion is influenced by the gaze, head pose, gesture, and facial appearance \cite{yan2020raf}. 
Besides, it is quite difficult to manually balance the two tasks and filter the less related FE samples.
%In the testing phase, merely the base net is utilized for AUs inference and estimation.

\subsection{Meta auxiliary learning}
Meta-learning (learning to learn) aims to understand and adapt learning itself on a higher level than merely acquiring subject knowledge \cite{lemke2015metalearning}. Meta-learning methods for the neural network have a long history \cite{thrun2012learning, bengio1990learning}, but have resurged in popularity recently \cite{finn2017model, li2019learning}.  
Currently, meta-learning studies have focussed on learning good weight initializations for few-shot learning \cite{finn2017model}, or learning to generate the optimal hyper-parameters/optimizers based on a meta net \cite{ravi2016optimization, andrychowicz2016learning}, or learning to generate useful auxiliary labels \cite{liu2019self}. For meta auxiliary learning,  Jaderberg et al. \cite{jaderberg2016reinforcement} incorporated the auxiliary tasks into the reinforcement learning framework in order to obtain ultimately higher performance for its agents. 
Kowshik et al. \cite{thopalli2019salt} formulated the subspace-based domain alignment as the auxiliary task to the primary one to achieve a generalizable classifier for both the source and the target domain.
Liu et al. \cite{liu2019self} proposed to automatically learn to generate appropriate auxiliary labels to benefit the primary task. Among them, the most related work to our proposed MAL is L2T-WW \cite{Jang2019learning}. L2T-WW benefits the primary task by automatically transferring knowledge from a pretrained model in a meta learning manner. It avoids storing the source dataset and is applicable where the source and target networks are heterogeneous. However, L2T-WW is not capable of adaptively balancing the multiple loss according to their unequal properties and behaviors. In addition, Without evaluation on the AU validation dataset, it is non-trivial to manually tune the optimal weights for the heterogeneous learning tasks.

\section{Method}
\begin{figure*}[htb]
	\centering
	\includegraphics[width=0.8\linewidth]{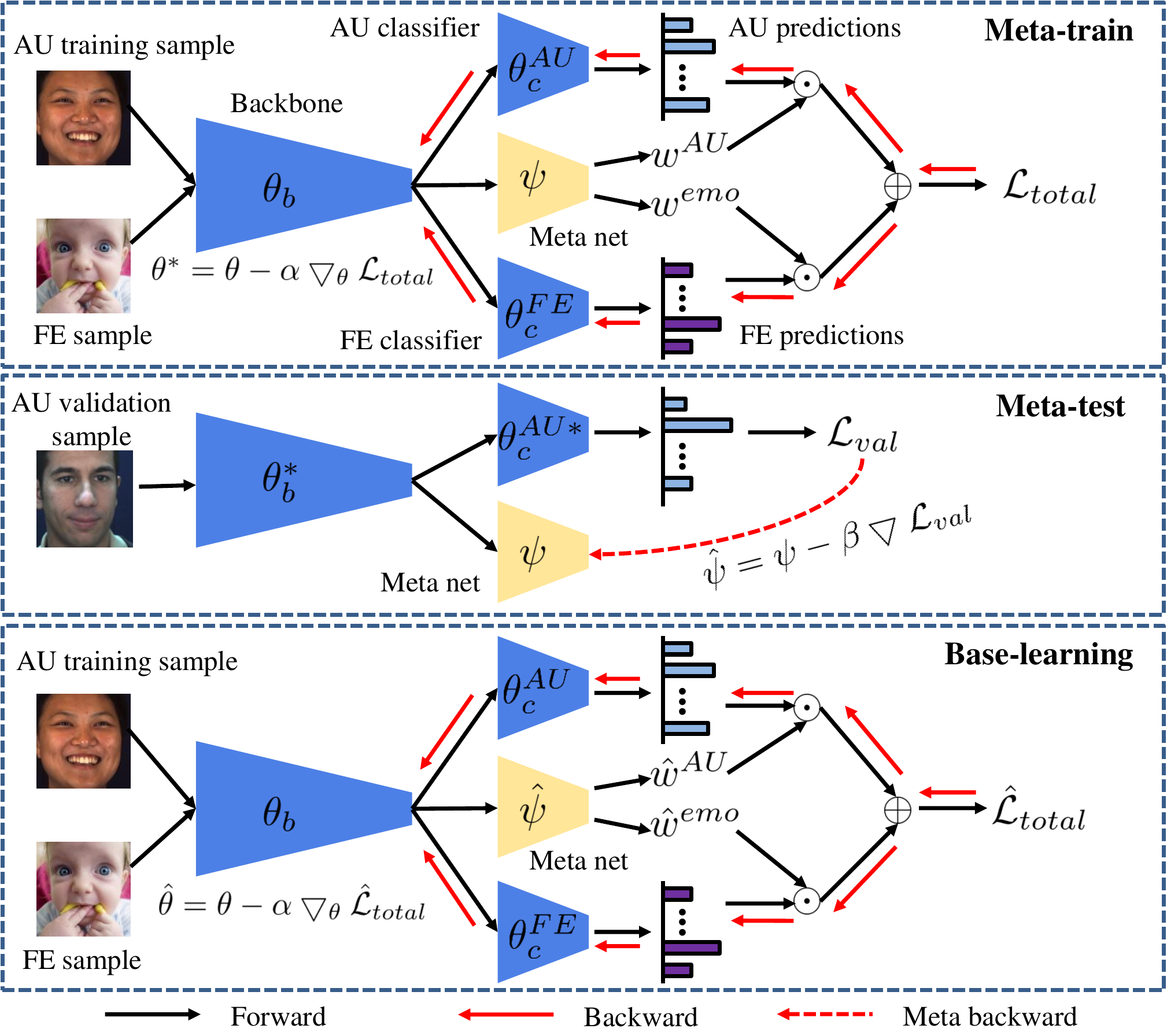}
	\caption{
		Illustration of the meta optimization pipeline of MAL. It consists of three stages: (i) Meta-train. (ii) Meta-test. (iii) Base-learning. 
		At the Meta-train stage, MAL updates the base net ($\theta \rightarrow \theta^{\ast}$) in a multi-task manner. 
		At the Meta-test stage, MAL evaluates the performance of the updated  base net, then \textit{meta}-update the meta net ($\psi \rightarrow \hat{\psi}$).
		At the Base-learning step, MAL learns the two tasks jointly with the updated sample weights ($\hat{w}^{AU}, \hat{w}^{FE}$). $\odot$ means element-wise multiplication. $\oplus$ means add operation.% Better viewed in color.
	}
	\label{fig:framework_overview}
	\vspace{-0.3cm}
\end{figure*}

In this section, we introduce the proposed Meta Auxiliary Learning method (MAL) in detail.
The overview of MAL is shown in Fig. \ref{fig:framework_overview}.
The core of MAL is a weighting function mapping from the image feature to the sample weight, and then iterating between weight recalculating and classifier updating. 
The network structure of MAL consists of a base net and a meta net. The two networks are optimized iteratively at each training iteration. Specially,  each training iteration of MAL consists of three steps: (i) the Meta-train that learns AU detection and FER in a multi-task manner. (ii) the Meta-test that \textit{meta-learns} the optimal weights for the training samples.
(iii) the Base-learning that learns AU detection and FER with the updated sample weights. 
By simulating the training-validation procedure at each training iteration, MAL ensures the optimization direction that minimizes the training loss leads to a decrease in the validation loss. The optimization trajectory of the base net is learned in a manner where gradients of training and validation samples are aligned \cite{li2018learning, ren2018learning}, thus the learned model can be better self-adapted to the testing AU samples.

% Our intuition is that when learning AU detection and FER in a multi-task manner, the FER images should be  In addition, humans perceive facial emotion in a personalized and culture-dependent manner,  the FER annotations may have different degrees of ambiguity\cite{li2017reliable, mollahosseini2017affectnet}.

\subsection{Overview}
Suppose in the training stage we have a AU training dataset $\mathcal{D}^{AU}_{tra} = \{(\mathcal{X}_{i}, z_{i}), 1 \leq i \leq N\}$, and a FE dataset $\mathcal{D}^{FE}_{tra} = \{(\mathcal{E}_{i}, y_{i}), 1 \leq i \leq M\}$.
Specially, we reserve a small unbiased validation data set $\mathcal{D}^{AU}_{val} = \{(\mathcal{V}_{i}, z_{i}), 1 \leq i \leq K\}$ that is sampled and excluded from the AU training dataset $\mathcal{D}^{AU}_{tra}$.
Among them, $\mathcal{X}_i, \mathcal{V}_i, \mathcal{E}_i$ denote the $i$th image in $\mathcal{D}^{AU}_{tra}$, $\mathcal{D}^{AU}_{val}$,  $\mathcal{D}^{FE}_{tra}$,  respectively. $N, K, M$ denote the total number of samples in $\mathcal{D}^{AU}_{tra}$, $\mathcal{D}^{AU}_{val}$, $\mathcal{D}^{FE}_{tra}$, and $K \ll N$.
$y_{i}$ is a one-hot vector representing the annotated category over $Q$ facial expression classes. $z_{i} \in \{0, 1\}$ represents annotation w.r.t the $i$th AU. 1 means the AU is active, 0 means inactive.

We exploit the multi-label sigmoid cross-entropy loss for AU detection. It is formulated as
\begin{equation}
	\mathcal{L}^{AU} = - \sum^{J}_{j} z^j \log\hat{z}^{j} + (1 - z^{j}) \log (1-\hat{z}^{j})
\end{equation}
where $J$ is the number of AUs. $z^j$ denotes the $j$th ground truth AU label of the input AU sample. $\hat{z}^j$ means the predicted AU score. 

For FER, we adopt the cross-entropy loss, which is formulated as:
\begin{equation}
	\mathcal{L}^{FE} = - \sum^{Q}_{q} y^q \log \hat{y}^q
\end{equation}
where $Q$ means the number of FE classes. $y^q$ and $\hat{y}^q$ denote the ground truth and predicted FE category, respectively. 

The conventional objective for the multi-task training is to minimize the combinated losses from all the individual task : $L_{train} = \mathcal{L}^{AU}(x,z) + \rho \mathcal{L}^{FE}(e,y)$, where $\rho$ represents the contribution of the FER task. Manually tuning of loss weights is tedious.
Instead,  MAL aims to automatically learn to assign the adaptive weights $w^{AU}_{i}$, $w^{FE}_{i}$ for each AU and FE sample with a meta-optimization objective, and minimize the loss:
\begin{equation}
	\label{equ:multi-task_pre}
	\mathcal{L}_{total} = \sum^{B}_{i=1} w^{AU}_{i} \mathcal{L}^{AU}_{i} + \sum^{B}_{i=1} w^{FE}_{i} \mathcal{L}^{FE}_{i},
\end{equation}
where $B$ means the mini-batch size. 

\subsection{Meta-optimization of MAL}
Fig. \ref{fig:framework_overview} illustrates the network structure and meta optimization pipeline of MAL.
The network structure of MAL consists of a base net and a meta net.
The base net consists of a backbone network followed by two parallel classifiers.  The two classifiers are respectively used for AU detection and FER.
We denote the backbone network as a function $f(\mathcal{X}_i)$ with parameters $\theta_{b}$ which takes as an input sample $\mathcal{X}_i$.
The parameters of the two classifiers are represented as  $\theta^{AU}_c$ and $\theta^{FE}_c$, respectively.
$\theta_{AU} = \{\theta_{b}, \theta^{AU}_c\}$ denotes the parameters that are related with the AU detection task, and likewise
$\theta_{FE} = \{\theta_{b}, \theta^{FE}_c\}$ means the parameters that are related with the FER task, thus the parameters in the base net can be represented as $\theta = \{\theta_{b}, \theta^{AU}_c, \theta^{FE}_c\}$.
The meta net receives the image feature $f(\mathcal{X}_i)$ as input and maps $f(\mathcal{X}_i)$ to a scalar weight $w$. 
We denote the meta net as a function $g(f(\mathcal{X}_i))$ with parameters $\psi$.

As illustrated in Fig. \ref{fig:framework_overview}, the meta optimization pipeline of of MAL consists of three stages: (i) \textbf{Meta-train}. (ii) \textbf{Meta-Test}. (iii) \textbf{Base-learning}. 
At each training iteration, MAL performs the three steps sequentially.
At the Meta-train stage, the base net takes as input a batch of AU and FE samples and computes the loss of each sample. The meta net estimates the initial weights $w^{AU}, w^{FE}$ for the AU, and FE samples. 
The losses of the two tasks are then scaled by their respective sample weights to update the base net in a multi-task manner ($\theta \rightarrow \theta^{\ast}$).
At the Meta-test stage, MAL takes as input a batch of AU samples from the validation set and evaluates the performance of the updated base net. MAL then performs a meta gradient descent step to update the meta net ($\psi \rightarrow \hat{\psi}$).
At the Base-learning step, MAL learns the two tasks jointly with the adaptive sample weights ($\hat{w}^{AU}, \hat{w}^{FE}$) to re-update the base net ($\theta \rightarrow \hat{\theta}$).

Below, we present the formulation and the details of the three stages.

\begin{algorithm}[htb]
	% 	\label{alg:mmtl_alg}
	\SetAlgoLined
	\KwIn{$\mathcal{D}^{AU}_{tra}$, $\mathcal{D}^{AU}_{val}$, $\mathcal{D}^{FE}_{tra}$}
	\textbf{Init}: hyperparameter $\alpha$, $\beta$, and batch size of $B$
	%\Kwln{$\alpha$}   % 
	
	\While{not converge}{
		\textbf{Meta-train:}\\
		{
			Sample a batch of AU samples $\mathcal{X}^{AU}_{tra}$ from $\mathcal{D}^{AU}_{tra}$\;
			Sample a batch of FE samples $\mathcal{E}^{FE}_{tra}$ from $\mathcal{D}^{FE}_{tra}$\;
			%$\mathcal{X}^{AU}_{tra}$ $\leftarrow$ SampleMiniBatch($\mathcal{D}^{AU}_{tra}$, B)\;
			%$\mathcal{E2}^{FE}_{tra}$ $\leftarrow$ SampleMiniBatch( $\mathcal{D}^{FE}_{tra}$, B)\;
			Compute the initial sample weights $w^{AU}_i$, $w^{FE}_i$  by Equ. \ref{equ:original_weight}\;
			Compute the initial multi-task loss by Equ. \ref{equ:multi-task}\;
			%$w^{AU}_i$ $\leftarrow$ Meta\_net\_forward($f(x_i), \psi)$\;
			%$w^{FE}_i$ $\leftarrow$ Meta\_net\_forward($f(e_i), \psi)$\;
			%$\mathcal{L}_{total} = \sum^{B}_{i=1} w^{AU}_{i} \mathcal{L}^{AU}_i + \sum^{B}_{i=1} w^{FE}_{i} \mathcal{L}^{FE}_{i}$\;
			Update the base net by:
			$\theta^{\ast}(\psi) = \theta -  \alpha \bigtriangledown_{\theta} \mathcal{L}_{total}$\;
		}
		\textbf{Meta-test:}\\
		{
			Sample a batch of AU validation samples $\mathcal{V}^{AU}_{val}$ from $\mathcal{D}^{AU}_{val}$\;
			%$\mathcal{X}^{AU}_{val}$ $\leftarrow$ SampleMiniBatch($\mathcal{D}^{AU}_{val}$, B)\;
			Update the meta net by meta backward by Equ.~\ref{equ:meta_update};
			%$\hat{\psi} = \psi - \beta \bigtriangledown_{\psi}(\bigtriangledown_{\theta^{\ast}_{AU}} \mathcal{L}_{val})$\;
		}
	
		\textbf{Base-learning:}\\
		{
			Update sample weight $\hat{w}^{AU}_i$, $\hat{w}^{FE}_i$  by Equ. \ref{equ:updated_weight}\;
			Compute the updated multi-task loss by Equ. \ref{equ:base_multitask}\;
			
			Re-update the base net by:
			$\hat{\theta} = \theta -  \alpha \bigtriangledown_{\theta} \hat{\mathcal{L}}_{total}$\;
			
		}
	}
	\caption{MAL for AU detection}
	\label{alg:mmtl_alg}
\end{algorithm}

\subsubsection{The Meta-train stage}
\label{sec:meta_train}
Fig. \ref{fig:framework_overview} (a) shows the Meta-train stage.  Given a batch of AU samples 
$\mathcal{X}^{AU}_{tra} = \{(\mathcal{X}_{i}, z_{i}), 1 \leq i \leq B \}$, and FE samples $\mathcal{E}^{FE}_{tra} = \{(\mathcal{E}_{i}, y_{i}), 1 \leq i \leq B \}$, the backbone network extracts their features as $f(\mathcal{X}_{i})$ and $f(\mathcal{E}_{i})$. With the encoded image features,  MAL obtains the per-sample weight by the meta net:
\begin{equation}
	%\begin{aligned}
	\begin{split}
		w^{AU}_{i} = g(f(\mathcal{X}_{i}); \psi), \\
		w^{FE}_{i} = g(f(\mathcal{E}_{i}); \psi).
	\end{split}
	%\end{aligned}
	\label{equ:original_weight}
\end{equation}

The losses of the AU detection and FER task are subsequently balanced by the estimated sample weights to obtain the multi-task objective:
\begin{equation}
	\label{equ:multi-task}
	%\begin{aligned}
	\mathcal{L}_{total} = \sum^{B}_{i=1} w^{AU}_{i} \mathcal{L}^{AU}_i(\theta_{AU}) + \sum^{B}_{i=1} w^{FE}_{i} \mathcal{L}^{FE}_{i}(\theta_{FE}).
	%\end{aligned}
\end{equation}

With the multi-task learning objective, MAL updates the parameters in the base net by a single stochastic gradient descent:
\begin{equation}
	\label{equ:theta_update_begin}
	\theta^{\ast}(\psi) = \theta -  \alpha \bigtriangledown_{\theta} \mathcal{L}_{total},
\end{equation}
where $\alpha$ means the learning rate. The parameters $\theta = \{\theta_{b}, \theta^{AU}_c, \theta^{FE}_c\}$ mean the parameters in the base net.

%Note that $w^{AU}_{i}$ and $w^{emo}_{i}$ are estimated through the meta net, the optimal example weight is learned base on the validation performance on $\mathcal{D}_{val}$.

\subsubsection{The Meta-test stage}
\label{sec:meta_test}

With the updated base net, MAL evaluates its one-step adaptation performance on the AU validation set $\mathcal{D}^{AU}_{val}$ to  meta-update the parameters $\psi$ in the meta net. 
Given a batch of AU validation samples  $\mathcal{V}^{AU}_{tra} = \{(\mathcal{V}_{i}, z_{i}), 1 \leq i \leq B \}$, MAL computes the loss $\mathcal{L}_{val}$ on $\mathcal{V}^{AU}_{tra}$, then updates the meta net by conducting a gradient descent step with regard to $\psi$:
\begin{equation}
	\hat{\psi} = \psi - \beta \bigtriangledown_{\psi} \mathcal{L}_{val}(\theta^{\ast}_{AU}),
	\label{equ:meta_update}
\end{equation}
where $\beta$ is the learning rate for the meta optimization \cite{finn2017model}, $\theta^{\ast}_{AU} = \{\theta^{\ast}_{b}, \theta^{AU\ast}_{c}\}$ means the parameters in the backbone network and the AU classifier. 
MAL takes a meta backward, i.e., a derivative over a derivative to update $\psi$. The meta-gradient $\bigtriangledown_{\psi} \mathcal{L}_{val}$ involves calculating a gradient through a gradient: 
\begin{equation}
	\bigtriangledown_{\psi} \mathcal{L}_{val} = \bigtriangledown_{\psi}(\bigtriangledown_{\theta^{\ast}_{AU}} \mathcal{L}_{val}).
\end{equation}

The second derivative operation was also adopted in several other meta-learning frameworks such as \cite{finn2017model, liu2019self, ren2018learning, Jang2019learning}.

\subsubsection{The Base-learning stage}
\label{sec:base_learning}
After the Meta-train and Meta-test stage, the meta net in MAL has been updated. With which MAL obtains the updated sample weights:
\begin{equation}
	\begin{split}
		\hat{w}^{AU}_{i} = g(f(\mathcal{X}_{i}); \hat{\psi}), \\
		\hat{w}^{FE}_{i} = g(f(\mathcal{E}_{i}); \hat{\psi}).
	\end{split}
	\label{equ:updated_weight}
\end{equation}

Then MAL computes the balanced multi-task loss on the current training batch samples:
\begin{equation}
	%\begin{aligned}
	\hat{\mathcal{L}}_{total} = \sum^{B}_{i=1} \hat{w}^{AU}_{i} \mathcal{L}^{AU}_{i}(\theta_{AU}) + \sum^{B}_{i=1} \hat{w}^{FE}_{i} \mathcal{L}^{FE}_{i}(\theta_{FE}).
	%\end{aligned}
	\label{equ:base_multitask}
\end{equation}

By minimizing $\hat{\mathcal{L}}_{total}$, MAL re-updates the parameters in the base net:
\begin{equation}
	\label{equ:theta_update_end}
	\hat{\theta} = \theta -  \alpha \bigtriangledown_{\theta} \hat{\mathcal{L}}_{total},
\end{equation}
where $\eta$ is the learning rate. $\hat{\theta}$ and $\hat{\psi}$  will be the initial parameters of the base net and the meta net in the next training iteration.

Different from the vanilla stochastic gradient descent for multi-task learning, MAL alternatively updates the base net parameters $\theta$ and the meta net parameters $\psi$ alternately. The full algorithm of the above three stages is outlined in Algorithm \ref{alg:mmtl_alg}, which can be implemented by popular deep learning frameworks such as pytorch \cite{paszke2019pytorch}.  

\section{Experiment}

\subsection{Implementation details}

\subsubsection{Network structures of MAL} We adopted a 34-layer ResNet \cite{he2016deep} as the backbone in the base net. The backbone network takes an RGB image in the size of $224 \times 224$ as the input and outputs a 512-dimensional image feature. 
The AU and FER classifiers both contain a fully connected layer that maps the image feature to the predicted probability.
The meta net contains a fully connected layer followed by a sigmoid layer. The parameters of the fully connected layer are initialized to zero, thus the initial weight for each sample is 0.5 at the beginning of training. The sigmoid layer in the meta net ensures the learned sample weight $w^{AU}_{i}$ and $w^{FE}_{i}$ range in  $[0, 1]$. This kind of net is known as a universal approximator for almost any continuous function\cite{csaji2001approximation} and can fitting a wide range of weighting functions.

\subsubsection{Training of MAL} The learning rates in MAL were set as: $\alpha = \beta =  0.001$. To match the predefined  training step size \cite{ren2018learning}, we normalized the average weights of AU and FE samples in each training batch so that they sum up to one. We calculated the average weight of AU/FE samples as $w^{AU}_{ave} = \sum_{i=1}^{B}{w^{AU}_{i}}$, $w^{FE}_{ave} = \sum_{i=1}^{B}{w^{FE}_{i}}$, where $B$ means the batch size.  Then the normalized sample weights are computed:$w^{AU}_{i} = \frac{w^{AU}_{i}}{w^{AU}_{ave} + w^{FE}_{ave}}$, $w^{FE}_{i} = \frac{w^{FE}_{i}}{w^{AU}_{ave} + w^{FE}_{ave}}$.
For the Meta-train and Base-learning stage, the batch sizes of AU and FE samples are set as 64. The input images are resized to a fixed size of $256 \times 256$ and randomly cropped into $224 \times 224$. Random horizontal flip is exploited for data augmentation.
For the Meta test stage, the batch size of AU validation data is set as 256.
We implemented all the experiments using PyTorch \cite{paszke2017automatic} on a Titan-X GPU with 12GB memory
We trained our proposed MAL for 30 epochs until convergence, it took approximately 54 hours to finish the training process.

\subsubsection{Evaluation of MAL} After the training process, we obtained the trained backbone network and the AU classifier for AU detection. FE samples were no longer required.
We adopted F1-score ($F1=\frac{2RP}{R+P}$) to evaluate the performance of MAL, where $R$ and $P$ denote recall and precision, respectively.  We additionally computed the average  F1-score over all AUs (ave) to measure the overall performance. We showed the AU detection results as $F1 \times 100$.

\subsubsection{Datasets} For AU detection, We adopted BP4D \cite{zhang2013high}, DISFA \cite{mavadati2013disfa}, and GFT \cite{girard2017sayette} datasets. 
Among them, BP4D is a spontaneous FACS dataset that consists of 328 videos for 41 subjects (18 males and 23 females). Each subject is involved in 8 sessions, and their spontaneous facial action units are captured with both 2D and 3D videos. 12 AUs are annotated for the 328
videos and there are about 142,000 frames with AU annotations of occurrence or absence.
 DISFA contains 27 participants from 12 females and 15 males. Each subject is asked to watch a 4-minute video to elicit facial AUs. The AUs are annotated with intensities from 0 to 5. In our experiments, We obtained nearly 130,000 AU-annotated images in the DISFA dataset by considering the images with intensities greater than 1 as active.  
GFT consists of 96 subjects in 32 three-person groups. Moderate out-of-plane head motion and occlusion are presented in the video frames, making AU detection in the GFT dataset challenging. 
For BP4D and DISFA dataset, we split the images into 3 folds in a subject-independent manner and conducted a 3-fold cross-validation. We adopted 12 AUs in BP4D and 8 AUs in DISFA dataset for evaluation. For GFT dataset, we utilized 10 AUs for evaluation. we followed the original train/test splits in \cite{girard2017sayette} and obtained nearly 108000 facial images for training and 24600 images for evaluation.

For the BP4D dataset and the GFT dataset, the pre-trained model based on ImageNet \cite{deng2009imagenet} dataset was used for initializing the backbone network in the base net.
For the DISFA dataset, we leveraged the model trained on BP4D to initialize the backbone network, following the same experimental setting of \cite{li2017eac, niu2019local}. 

To obtain the AU validation dataset, we randomly reserved approximately 2\% facial images from each AU training dataset. The selected images are uniformly sampled from each subject to increase the diversity. Considering  the reserved validation AU datasets are highly imbalanced, we reweighed samples from the under-represented categories inversely proportionally to the class frequencies,  aiming to alleviate the class imbalance among different AUs.

For facial expression recognition, we leveraged the RAF-DB dataset \cite{li2017reliable} for training.  RAF-DB is a widely used in-the-wild FE dataset that contains about 30,000 facial images annotated basic or compound emotions by 40 trained human coders. In our experiment,  only images with seven basic emotions were adopted, including about 12,300 images.

\begin{table*}[htb]
	\caption{F1 score on the BP4D dataset. \textbf{Bold} denotes the best among the state-of-the-art multi-task and meta auxiliary learning methods.  \underline{Underline} denotes the best among various state-of-the-art AU detection methods. * means the values are reported in the original papers.}
	\centering
	\small
	\label{tab:bp4d_cross_dataset}
	\begin{tabular}{|c|c|c|c|c|c|c|c|c|c|c|c|c|c|}
		\hline
		Methods & AU1 & AU2 & AU4 & AU6 & AU7 & AU10 & AU12 & AU14 & AU15 & AU17 & AU23 & AU24 & ave \\   
		\hline
		\hline
		%LSVM* & 23.2 & 22.8 & 23.1 & 27.2 & 47.1 & 77.2 & 63.7 & 64.3 & 18.4 & 33.0 & 19.4 & 20.7 & 35.3 \\
		DRML \cite{zhao2016deep}*  & 36.4 & 41.8 & 43.0 & 55.0 & 67.0 & 66.3 & 65.8 & 54.1 & 33.2 & 48.0 & 31.7 & 30.0 & 48.3 \\
		EAC-Net \cite{li2017eac}* &  39.0 & 35.2 & 48.6 & 76.1 & 72.9 & 81.9 & 86.2 & 58.8 & 37.5 & 59.1 & 35.9 & 35.8 & 55.9 \\
		ROI \cite{li2017action}*  &  36.2 & 31.6 & 43.4 & 77.1 & 73.7 & \underline{85.0} & 87.0 & 62.6 & 45.7 & 58.0 & 38.3 & 37.4 & 56.4 \\
		JAA-Net \cite{shao2018deep}*  & 47.2 & 44.0 & 54.9 & \underline{77.5} & 74.6 & 84.0 & 86.9 & 61.9 & 43.6 & 60.3 & 42.7 & 41.9 & 60.0 \\
		DSIN \cite{corneanu2018deep}* & \underline{51.7} & 40.4 & \underline{56.0} & 76.1 & 73.5 & 79.9 & 85.4 & 62.7 & 37.3 & 62.9 & 38.8 & 41.6 & 58.9 \\
		LP-Net \cite{niu2019local} * & 43.4 & 38.0 & 54.2 & 77.1 & 76.7 & 83.8 & 87.2 & 63.3 & 45.3 & 60.5 & \underline{48.1} & \underline{54.2} & 61.0 \\
		% TCAE* &  43.1 & 32.2 & 44.4 & 75.1 & 70.5 & 80.8 & 85.5 & 61.8 & 34.7 & 58.5 & 37.2 & 48.7 & 56.1\\
		TAE \cite{li2020learning}* & 47.0 & \underline{45.9} & 50.9 & 74.7 & 72.0 & 82.4 & 85.6 & 62.3 & 48.1 & 62.3 & 45.9 & 46.3 & 60.3\\
		gBN \cite{cui2020knowledge}* & - & - & - & - & - & - & - & - & - & - & - & - & 57.0 \\
		SRERL \cite{li2019semantic}* & 46.9 & 45.3 & \underline{55.6} & 77.1 & \underline{78.4} & 83.5 & \underline{87.6} & \underline{63.9} & \underline{52.2} & \underline{63.9} & 47.1 & 53.3 & \underline{62.9} \\
		\hline
		\hline
		MTAN \cite{liu2019end} &  43.4 & 45.1 & 50.9 & 76.0 & 75.0 & \textbf{83.1} & 86.8 & 60.9 & 47.4 & 62.3 & 37.9 & 50.5 & 59.9\\
		L2T-WW \cite{Jang2019learning} &  41.6 & \textbf{49.5} & 48.3 & 74.1 & 77.2 & 81.6 & 83.8 & 60.0 & 46.2 & \textbf{62.9} & 41.8 & \textbf{52.4} & 60.0\\
		STL  &  46.7 & 44.0 & 49.7 & 73.0 & 74.5 & 81.4 & 85.5 & 61.6 & 48.2 & 57.4 & 44.3 & 44.5 & 59.2\\
		MTL  &  45.9 & 41.6 & 50.8 & 75.2 & 74.3 & 82.6 & 87.0 & 59.8 & 47.1 & 61.3 & 44.6 & 45.5 & 59.7\\
		\textbf{MAL(Ours)} & \textbf{47.9} & \textbf{49.5} & \textbf{52.1} & \textbf{77.6} & \textbf{77.8} & 82.8 & \textbf{88.3} & \textbf{66.4} & \textbf{49.7} & 59.7 & \textbf{45.2} & 48.5 & \textbf{62.2} \\
		\hline
	\end{tabular}
\end{table*}

\begin{table*}[htb]
	\centering
	\small
	\caption{F1 score on the DISFA dataset. \textbf{Bold} denotes the best among the state-of-the-art multi-task and meta auxiliary learning methods.  \underline{Underline} denotes the best among various state-of-the-art AU detection methods. * means the values are reported in the original papers.}
	\label{tab:disfa_cross_dataset}
	\begin{tabular}{|c|c|c|c|c|c|c|c|c|c|}
		\hline
		Methods  & AU1 & AU2 & AU4 & AU6 & AU9 & AU12 & AU25 & AU26 & ave \\
		\hline
		\hline
		%LSVM* & 10.8 & 10.0 & 21.8 & 15.7 & 11.5 & 70.4 & 12.0 & 22.1 & 21.8 \\
		DRML \cite{zhao2016deep}*  &  17.3 & 17.7 & 37.4 & 29.0 & 10.7 & 37.7 & 38.5 & 20.1 & 26.7 \\
		EAC-Net \cite{li2017eac}*  &  41.5 & 26.4 & 66.4 & 50.7 & \underline{80.5} & \underline{89.3} & 88.9 & 15.6 & 48.5 \\
		JAA-Net \cite{shao2018deep}*  & 43.7 & 46.2 & 56.0 & 41.4 & 44.7 & 69.6 & 88.3 & 58.4 & 56.0 \\
		OFS-CNN \cite{han2018optimizing}* & 43.7 & 40.0 & 67.2 & 59.0 & 49.7 & 75.8 & 72.4 & 54.8 & 51.4 \\
		DSIN \cite{corneanu2018deep}*   & 42.4 & 39.0 & 68.4 & 28.6 & 46.8 & 70.8 & 90.4 & 42.2 & 53.6  \\
		LP-Net \cite{niu2019local} * & 29.9 & 24.7 & \underline{72.7} & 46.8 & 49.6 & 72.9 & \underline{93.8} & \underline{65.0} & \underline{56.9} \\
		% TCAE*  &  15.1 & 15.2 & 50.5 & \underline{48.7} & 23.3 & 72.1 & 82.1 & 52.9 & 45.0 \\
		TAE \cite{li2020learning}* & 21.4 & 19.6 & 64.5 & 46.8 & 44.0 & 73.2 & 85.1 & 55.3 & 51.5 \\
		SRERL \cite{li2019semantic}* & \underline{45.7} & \underline{47.8} & 59.6 & 47.1 & 45.6 & 73.5 & 84.3 & 43.6 & 55.9 \\
		\hline
		\hline
		MTAN  \cite{liu2019end} &  35.7 & 32.3 & 64.3 & 56.5 & 35.5 & 72.5 & 87.9 & \textbf{62.6} & 55.9 \\
		L2T-WW \cite{Jang2019learning} &  32.3 & 27.7 & 62.2 & 47.6 & 36.2 & 73.8 & 85.4 & 55.1 & 52.6 \\
		STL  &  38.0 & 33.1 & 51.8 & 46.2 & 34.2 & 65.4 & 85.4 & 56.9 & 51.3 \\
		MTL  &  40.6 & 35.2 & 66.9 & \textbf{48.6} & 35.3 & 72.5 & 88.0 & 53.1 & 55.0 \\
		\textbf{MAL(Ours)} & \textbf{43.8} & \textbf{39.3} & \textbf{68.9} & 47.4 & \textbf{48.6} & \textbf{72.7} & \textbf{90.6} & 52.6 & \textbf{58.0} \\
		\hline
	\end{tabular}
\end{table*}

\begin{table*}[htb]
	\centering
	\small
	\caption{F1 score on the GFT dataset. \textbf{Bold} denotes the best among the state-of-the-art multi-task and meta auxiliary learning methods.  \underline{Underline} denotes the best among various state-of-the-art AU detection methods. * means the values are reported in the original papers.}
	\label{tab:gft_cross_dataset}
	\begin{tabular}{|c|c|c|c|c|c|c|c|c|c|c|c|}
		\hline
		Methods & AU1 & AU2 & AU4 & AU6 & AU10 & AU12 & AU14 & AU15 & AU23 & AU24 & ave \\
		\hline
		\hline
		AlexNet \cite{girard2017sayette} *  &  44 & 46 & 2 & 73 & 72 & 82 & 5 & 19 & 43 & 42 & 42.8  \\
		ResNet-50 \cite{he2016deep} & 23.5 & 37.8 & 3.5 & \underline{79.1} & 70.1 & \underline{82.1} & \underline{20.9} & 11.7 & 49.1 & 40.3 & 41.8 \\
		% TCAE* &  43.9 & \underline{49.5} & 6.3 & 71.0 &  \underline{76.2} & 79.5 & 10.7 & 28.5 & 34.5 & 41.7 & 44.2 \\
		TAE  \cite{li2020learning}*  &  \underline{46.3} & 48.8 & 13.4 & 76.7 & 74.8 & 81.8 & 19.9 & \underline{42.3} & \underline{50.6} & \underline{50.0} & 50.5 \\
		\hline
		\hline
		MTAN \cite{liu2019end} &  44.7 & 49.6 & 27.0 & 79.0 & \textbf{78.9} & 83.1 & 32.5 & \textbf{49.2} & 53.8 & 51.2 & 54.9 \\
		L2T-WW \cite{Jang2019learning} &  40.8 & 46.0 & 42.7 & \textbf{83.9} & 77.1 & 83.1 & 27.1 & 42.0 & 53.9 & \textbf{53.7} & 55.0 \\
		STL &  43.2 & 48.2 & 25.9 & 74 & 73.7 & 83.5 & 37.5 & 44.0 & 51.7 & 48.4 & 53.0 \\
		MTL &  30.7 & 54.7 & 41.1 & 82.7 & 75.4 & 84.2 & 40.5 & 46.5 & 53.2 & 50.6 & 56.0 \\
		\textbf{MAL(Ours)} & \textbf{52.4} & \textbf{57.0} & \textbf{54.1} & 74.5 & 78.0 & \textbf{84.9} & \textbf{43.1} & 47.7 & \textbf{54.4} & 51.9 & \textbf{59.8} \\
		\hline
	\end{tabular}
\end{table*}

\subsection{Comparison with the state-of-the-art methods}
We compared MAL with other state-of-the-art AU detection methods, multi-task learning methods, and meta auxiliary learning methods. Table \ref{tab:bp4d_cross_dataset}, \ref{tab:disfa_cross_dataset}, \ref{tab:gft_cross_dataset} illustrate the F1-score of the methods on BP4D, DISFA, and GFT datasets.

\subsubsection{Comparison with other AU detection methods} We compare MAL with the state-of-the-art AU detection methods, including DRML \cite{zhao2016deep}, Enhancing and Cropping Net (EAC-Net) \cite{li2017eac}, Region Adaptation and Multi-Label Learning (ROI) \cite{li2017action}, JAA-Net \cite{shao2018deep}, Local Relationship Learning (LP-Net) \cite{niu2019local}, OFS-CNN\cite{han2018optimizing}, Deep Structure Inference Network (DSIN) \cite{corneanu2018deep}, TCAE\cite{li2019self}, TAE \cite{li2020learning}, gBN \cite{cui2020knowledge}, SRERL \cite{li2019semantic}. 
Among them, most of the AU methods (including ROI \cite{li2017action},  JAA-Net \cite{shao2018deep}, DSIN \cite{corneanu2018deep}, SRERL \cite{li2019semantic}) extract the local facial regions with manually defined AU centers and learn the AU-specific representations with exclusive CNN branches.
TAE \cite{li2020learning} exploits VoxCeleb1 [56] and VoxCeleb2 [57] datasets that consist of around 7,000 subjects to learn the AU-discriminative representation in a self-supervised mode. 
gBN \cite{cui2020knowledge} integrates the facial expression results to augment the AU detection model.
These methods are evaluated under the same protocol on the three FACS datasets.

Table \ref{tab:bp4d_cross_dataset}, \ref{tab:disfa_cross_dataset}, \ref{tab:gft_cross_dataset} show that MAL is comparable with other state-of-the-art AU detection methods on BP4D dataset. 
On BP4D dataset, our proposed MAL is comparable with SRERL whose network structure consists of a VGG19 backbone followed by 24 sub-branches, as well as a Gated Gated Graph Neural Network (GGNN) for explicit AU relation modeling. MAL outperforms DRML \cite{zhao2016deep}, EAC-Net \cite{li2017eac}, ROI \cite{li2017action}, JAA-Net \cite{shao2018deep}, DSIN  \cite{corneanu2018deep}, LP-Net \cite{niu2019local}, TAE \cite{li2020learning}, gBN \cite{cui2020knowledge} in the average F1 score. MAL additionally outperforms other state-of-the-art AU detection methods on AU1 (inner brow raiser), AU2 (outer brow raiser), AU6 (cheek raiser), AU12  (lip corner puller), AU14 (dimpler).
It indicates the superiority of our proposed meta auxiliary method as MAL does not depend on facial landmarks to extract facial local regions and learns the AU-discriminative representation in a data-driven manner.

On the DISFA dataset, our proposed MAL outperforms all the compared state-of-the-art AU detection methods. It is because DISFA is a challenging dataset as it merely consists of 27 subjects. This data scarcity significantly limits the upper bound of the supervised methods that merely train the AU detection model with merely the facial images in the DISFA dataset. As reported in \cite{girard2015much}, variation in the number of subjects rather than the number of frames per subject yields the most efficient performance. It means the number of the training subjects plays a key role in the final performance of the AU detection model. Our proposed MAL incorporates diverse subjects in the training process and optimizes the AU detection model towards enhancing the performance on the AU validation dataset. The FE samples usually consist of the corresponding AUs activated, e.g., activated AU1 (inner brow raiser) often appears in FE images labeled with \textit{fear}, activated AU4 (brow lowerer) often appears in FE images labeled with \textit{anger},  activated AU6  (cheek raiser), AU12  (lip corner puller) often appears in FE images labeled with \textit{happy}, AU1 (inner brow raiser), AU15 (lip corner depressor) often appears in FE images labeled with \textit{sad}, activated  AU2 (outer brow raiser), AU26 (jaw drop) often appears in FE images labeled with \textit{surprise}, thus the FE images greatly augment the diversity of the training samples for training.

On the GFT dataset, our proposed MAL outperforms other state-of-the-art AU detection methods with considerable improvements. MAL obtains the best average F1 score and shows its success on AU1 (inner brow raiser), AU4 (brow lower), AU6(cheek raiser), AU12 (lip corner puller), AU14 (dimpler), AU15 (lip corner depressor), AU23 (lip tightener). MAL outperforms the self-supervised AU detection method TAE \cite{} with 9.3\% percentages.
It is worth mentioning that the state-of-the-art AU detection methods usually learn region-specific AU representation and inevitably depend on manually designed local facial regions in grids \cite{zhao2016deep}, or around facial landmarks \cite{li2017eac, li2017action, shao2018deep, corneanu2018deep, li2019semantic}, which inevitably increase the model complexity as well as the inference time. 
For the facial images that show large head pose such as GFT, the cropped patches according to the facial landmarks will suffer from misalignments and the patches would not be semantically aligned for different facial images.
Our proposed MAL merely adopts a vanilla 34-layer ResNet that does not rely on facial landmarks for facial region decomposition and does not introduce extra computation burden during the inference phase.  With the iterative optimization algorithm, MAL is capable of adaptatively transferring knowledge from a large amount of FE samples to boost the AU detection task, thus is also capable of mitigating the data scarcity issue and enhancing the generalization capability of the learned AU detection model. 

\begin{figure*}[htb]
	\centering
	\begin{subfigure}
		\centering
		\includegraphics[width=0.6\linewidth]{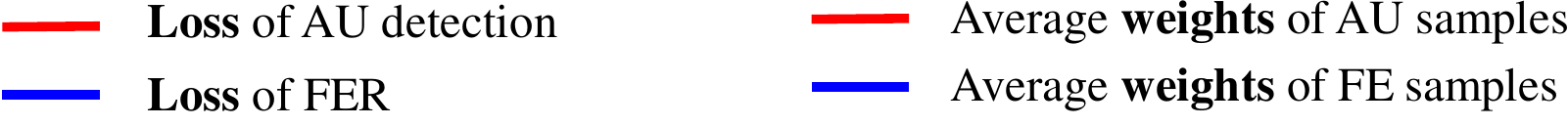}
		%\label{fig:retrieval-a}
	\end{subfigure}%
	\hspace{.2in}
	\begin{subfigure}
		\centering
		\includegraphics[width=0.8\linewidth]{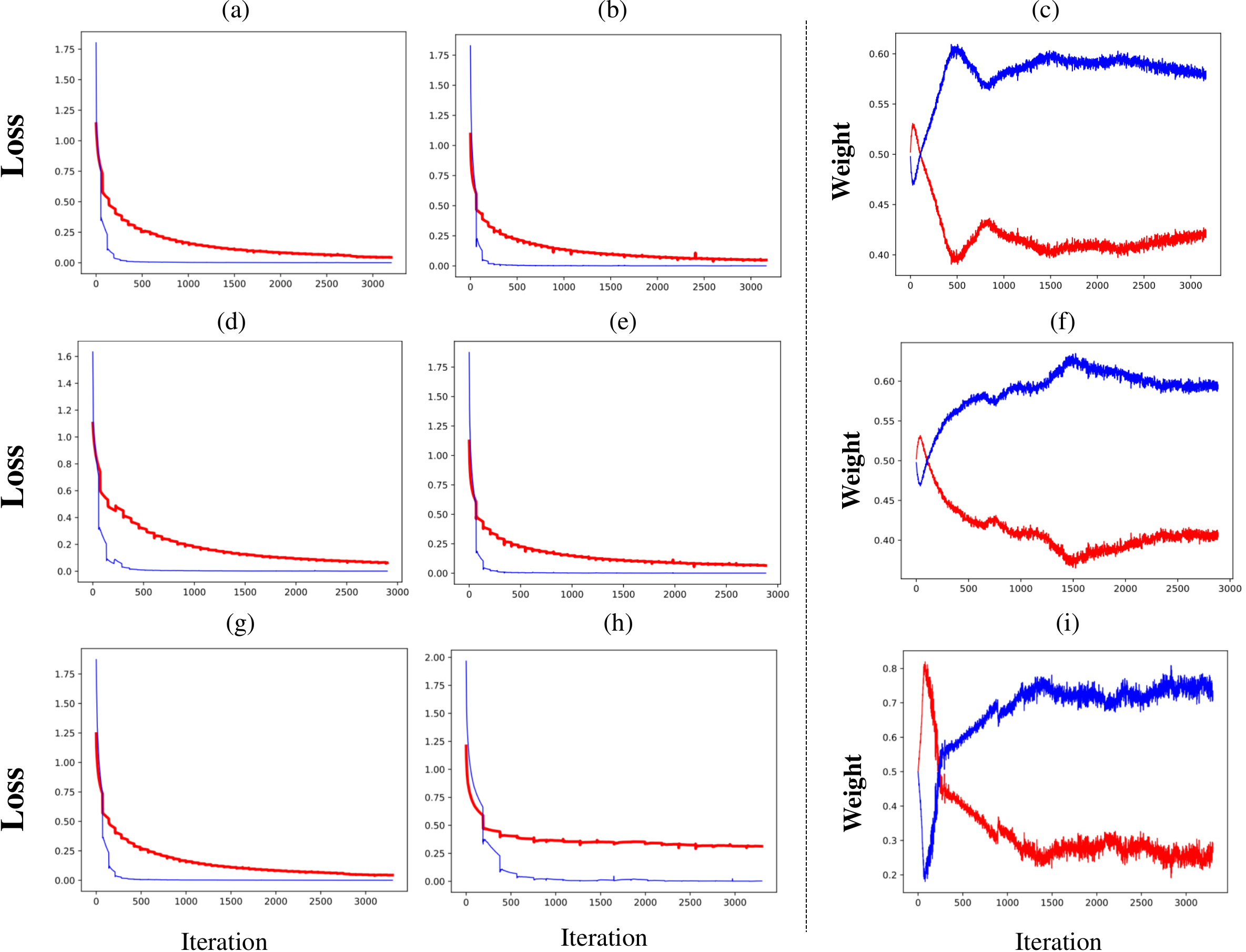}
		%\caption{1a}
		%\label{fig:retrieval-b}
	\end{subfigure}%
	\caption{
		Iteration-loss (a) and iteration-weight (b) curves of the training process on the BP4D dataset.
		The left and middle columns show the iteration-loss curves of MTL and our proposed MAL on each fold in the BP4D dataset, respectively. The right column shows the iteration-weight curves of our proposed MAL. Note that the average sample weights in (c),(f),(i) are automatically learned by our proposed MAL.
	}
	\label{fig:bp4d_weight_loss}
\end{figure*}

\subsubsection{Comparison with the multi-task and meta auxiliary learning methods}  We compare our proposed MAL with the state-of-the-art multi-task and meta auxiliary learning method, including Multi-Task Attention Network (MTAN) \cite{liu2019end}, L2T-WW. 
Among them, MTAN is the stage-of-the-art multi-task learning approach, it consists of a single shared network that contains a global feature pool and a soft-attention module for every single task. MTAN leverages the shared network to learns task-shared features and exploits multiple attention networks to learn the task-specific features. We re-trained the MTAN model following the same settings in MAL using the released codes.
 L2T-WW \cite{Jang2019learning} falls into the transfer learning paradigm that exploits meta-learning to automatically learn what knowledge to transfer from the source network to where in the target network.
 To re-implement L2T-WW, we trained a 34-layer ResNet on the RAF-DB dataset to serve as the source model, which was used to transfer knowledge so as to boost the target AU detection task in a meta-learning manner.

As shown in table \ref{tab:bp4d_cross_dataset}, \ref{tab:disfa_cross_dataset}, \ref{tab:gft_cross_dataset}, our proposed MAL outperforms MTAN with 2.3\%, 2.1\%, 4.9\% improvements in the average F1 score on BP4D, DISFA, GFT datasets, respectively.
On the evaluated three datasets, our proposed MAL outperforms MTAN on most of the AUs presented in table \ref{tab:bp4d_cross_dataset}, \ref{tab:disfa_cross_dataset}, \ref{tab:gft_cross_dataset}, including AU1 (inner brow raiser), AU2 (outer brow raiser), AU4 (brow lowerer), AU6(cheek raiser), AU7 (lid tightener), AU12 (lip corner puller), AU14 (dimpler), AU23 (lip tightener), AU24 (lips part).  .
The benefits of our proposed MAL over MTAN indicate the benefits of the proposed meta optimization algorithm that supervise the AU detection model towards enhancing the performance on the reserved AU validation dataset.
It also suggests the feasibility of the adaptatively learned sample weights. By meta-learning the weights for each training sample in the FER and AU detection tasks, MAL automatically balances the two tasks during the training process and is capable of filtering the less related FE samples. 

Compare with the state-of-the-art meta auxiliary learning method L2T-WW \cite{Jang2019learning}, our proposed MAL obtains 2.2\%, 5.4\%, 4.8\% improvements in the average F1 score on BP4D, DISFA, and GFT dataset, respectively.
MAL shows its success on various AUs such as AU1 (inner brow raiser), AU2 (outer brow raiser), AU4 (brow lowerer), AU7 (lid tightener), AU10 (upper lip raiser), AU14 (dimpler), AU15 (lip corner depressor), AU23 (lip tightener).
The improvements over L2T-WW  show the benefits of MAL is two-fold: 1) MAL transfer knowledge from FE datasets by adaptatively weighing each FE sample in a fine-grained manner. 2) MAL optimizes the AU detection model towards enhancing the performance of the AU validation dataset. By selectively transferring  knowledge from the informative FE samples, our proposed MAL obtains consistent improvements in both the average F1 score and on individual facial AU.

\begin{figure}[htb]
	\centering
	\includegraphics[width=0.9\linewidth]{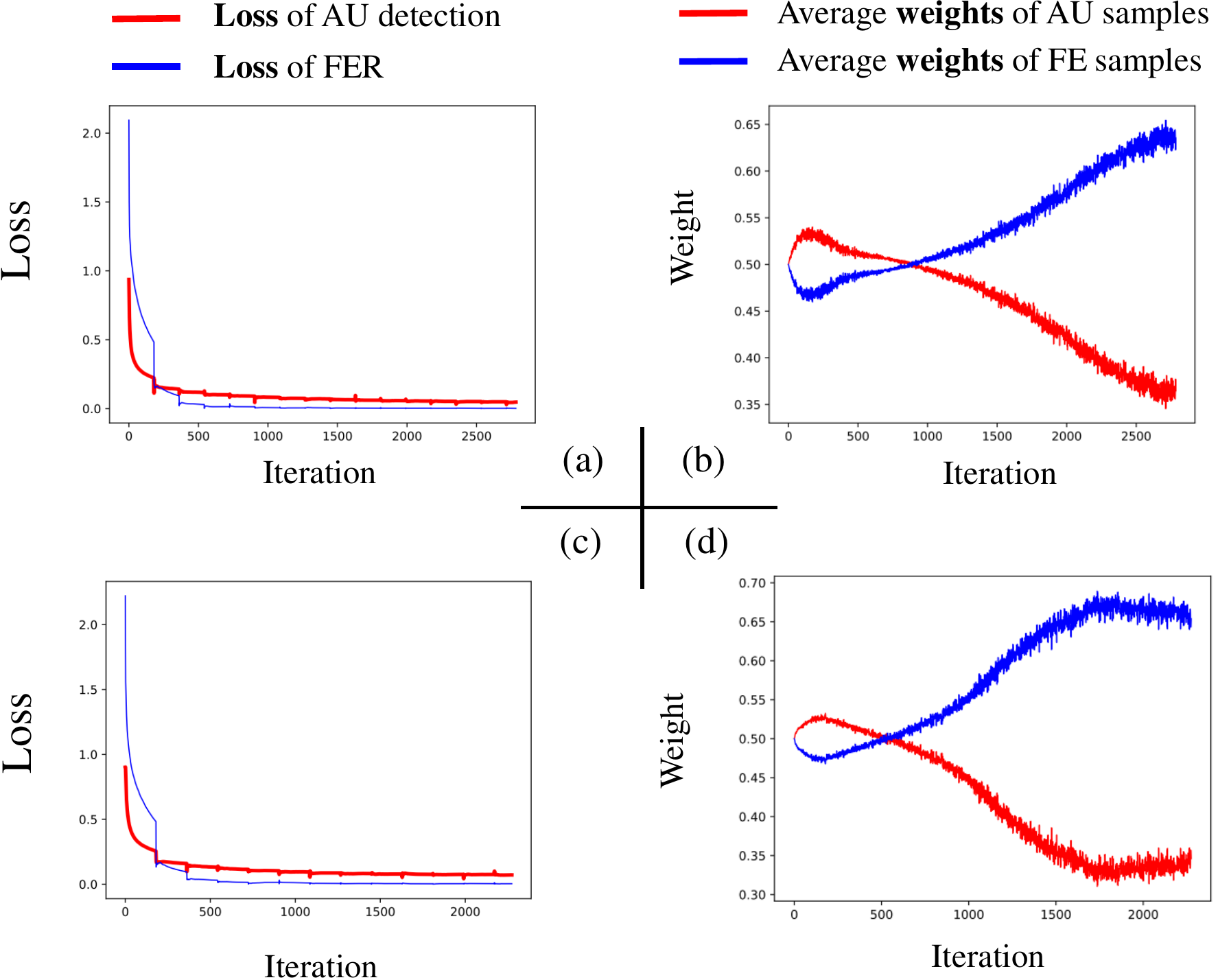}
	\caption{
		Iteration-loss ((a), (c)) and iteration-weight ((b),(d)) curves of the training process on DISFA dataset.
		The average sample weights in (b) and (d) are automatically learned by MAL.
	}
	\label{fig:disfa_weight_loss}
\end{figure}

\begin{figure*}[htb]
	\centering
	\includegraphics[width=0.8\linewidth]{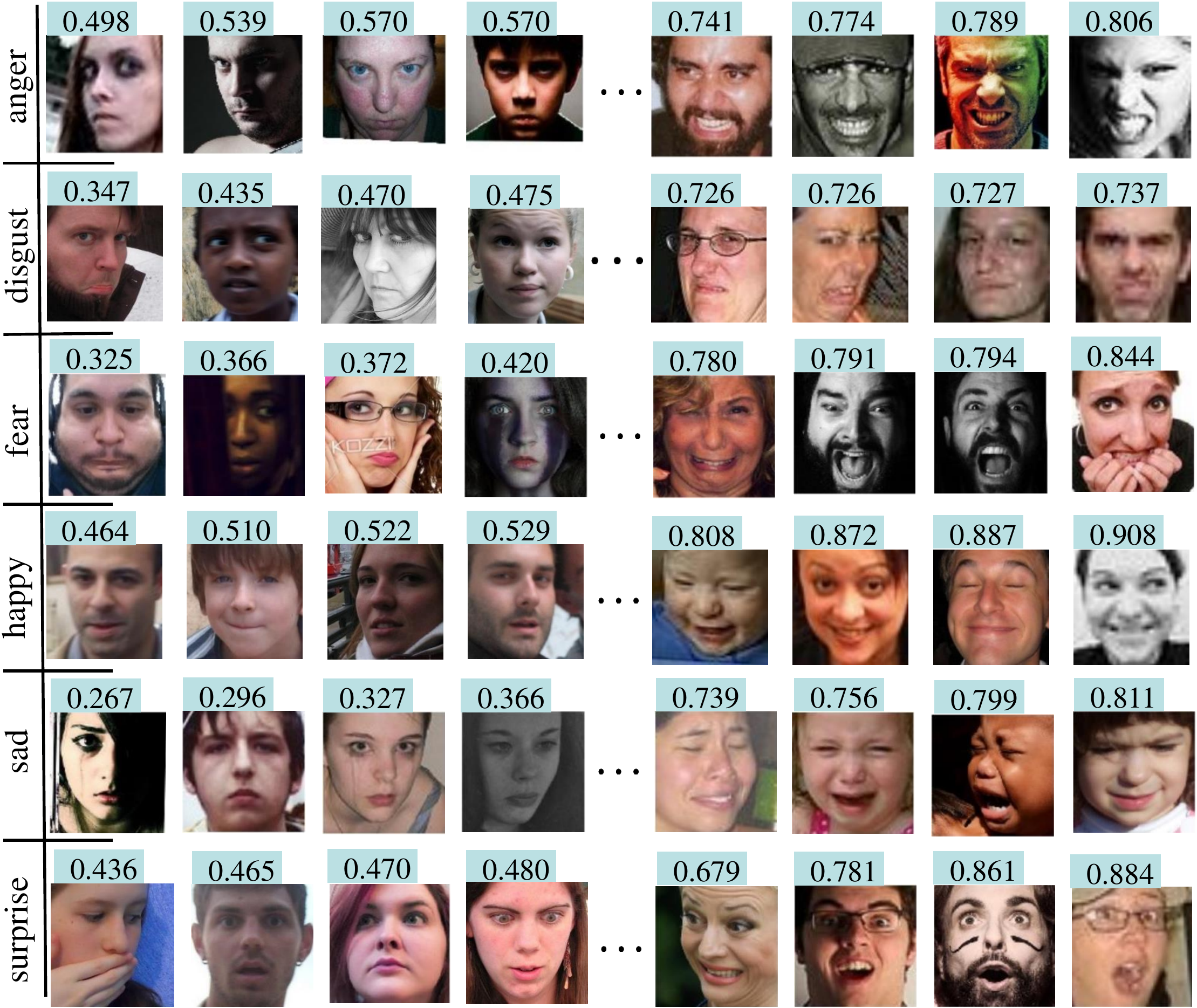}
	\caption{
		Illustration of the learned FE sample weights. The weight of each sample is displayed in the top-left rectangle.
		Each row shows a sequence of FE samples of the same category, arranged in the ascending order of the learned sample weights. 
		It is clear that the FE samples with smaller weight are usually distinguished by their gazes, head poses and gestures \cite{yan2020raf}, not by their facial AUs .
	}
	\label{fig:img_weight_vis}
\end{figure*}

\begin{figure*}[htb]
	\centering
	\includegraphics[width=0.9\linewidth]{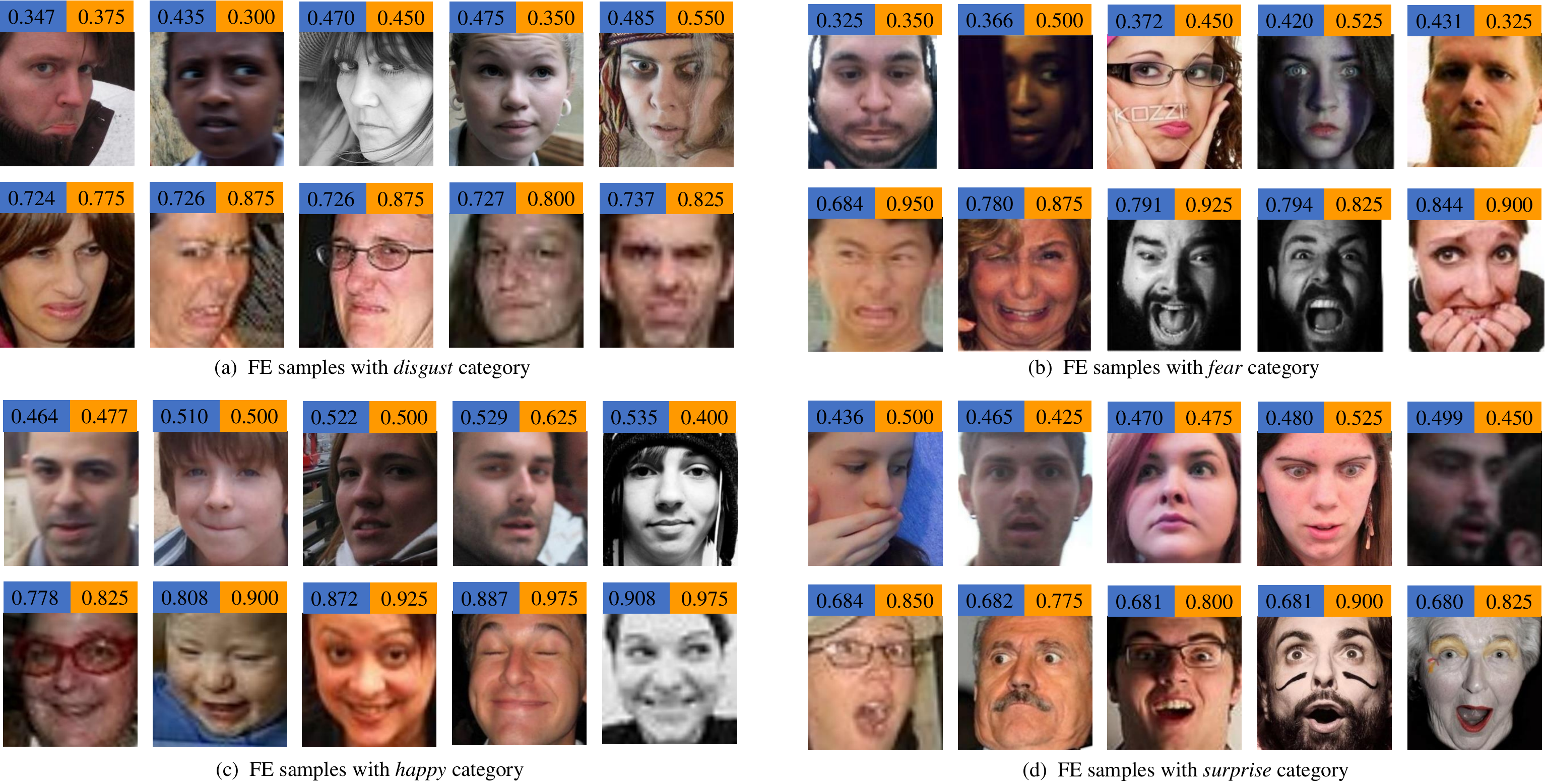}
	\caption{
		Illustration of the learned FE sample weights and the corresponding consistency value. 
		The larger the consistency value, the lower the uncertainty.
		It is clear the FE samples with low weights usually have small consistency values, their facial expressions are mainly categorized by their gazes, head poses, or gestures. The FE samples with high consistency value usually have the corresponding AUs activated. 
		}
	\label{fig:image_weight_consistency}
\end{figure*}

\subsection{Ablation study}
We conducted a quantitative evaluation to investigate the performance of the conventional multi-task learning method (MTL), and the single-task learning method (STL).  We implemented the STL and  MTL to compare their peformance with our proposed MAL on BP4D, DISFA, GFT datasets, as illustrated in in table \ref{tab:bp4d_cross_dataset}, \ref{tab:disfa_cross_dataset}, \ref{tab:gft_cross_dataset}.
For STL, the network structure merely consists of the backbone network and the AU classifier in MAL. 
For MTL, the network structure is the same as that of the base net in MAL. We trained the MTL model with the AU detection and FER task, the loss weights of the facial expression recognition and AU detection tasks are set as $1:1$.

As illustrated in table \ref{tab:bp4d_cross_dataset}, \ref{tab:disfa_cross_dataset}, \ref{tab:gft_cross_dataset}, MTL outperfroms STL on most AUs on DISFA and GFT datasets. 
 The improvements of MTL over STL are in line with the observation in \cite{pons2018multi, wang2019multi} and indicates the effectiveness of the multi-task method that incorporates  additional FE samples for training.
 The improvements are reasonable as the additional FE training data incorporates a large number of subjects which play a key role in the final performance of the AU detection model \cite{girard2015much}.
 On the BP4D dataset, STL is comparable with MTL. It also suggests that the considerable enhancement of one single task in MTL is not always guaranteed. For MTL, different  weighting strategies tend to work best for different tasks or datasets \cite{misra2016cross}. Besides, it is not capable of adaptatively selecting the informative FE samples to boost AU detection. Our proposed mitigates this issue and obtains consistent improvements over STL, as illustrated in table \ref{tab:bp4d_cross_dataset}, \ref{tab:disfa_cross_dataset}, \ref{tab:gft_cross_dataset}.
 MAL shows its superiority in each AU in the presented tables.
 Compare with MTL, our proposed MAL obtains 2.5\%, 3.0\%, 3.8\% improvements in the average F1 score.
It indicates our proposed MAL is capable of adaptatively transfer knowledge from the FE samples, the learned sample weights are effective to suppress the influence of the negative transfer in the conventional multi-task learning paradigm. We analyze the learned FE sample weights in the next section.

\subsection{Analysis}
We visualized the learned sample weights during the training process to investigate how MAL decreases the influence of negative transfer, i.e.,  how the two tasks are adaptatively balanced, and how the less related FE samples are suppressed?

We visualize the iteration-loss and iteration-weight curves during the training process in Fig. \ref{fig:bp4d_weight_loss}. For each input batch of AU and FE samples, we calculated their respective average weights and visualized the loss as well as average weights for every 20 iterations. For a fair comparison, we additionally visualize the iteration-loss curve of the conventional multi-task learning (MTL) method.
In Fig.~\ref{fig:bp4d_weight_loss}, the left and middle columns show the iteration-loss curve of MTL and our proposed MAL on the BP4D dataset, respectively. The right column shows the iteration-weight curves of our proposed MAL.
In Fig.~\ref{fig:disfa_weight_loss}, the left column shows the iteration-loss curve of MTL and our proposed MAL on the DISFA dataset, respectively. The right column shows the iteration-weight curves of our proposed MAL.

As can be seen in Fig.~\ref{fig:bp4d_weight_loss} (a),(d),(g), the loss of FER decreases faster than AU detection as the training progresses in MTL, it suggests that the FER task is relatively easy to optimize and dominates the training process in MTL.  The similar phenomenon can be observed in Fig.~\ref{fig:disfa_weight_loss} and Fig.~\ref{fig:bp4d_weight_loss} (b),(e),(h). 
With the increase of the training iterations, the loss of the auxiliary FER task dramatically decreases. It means the benefits of the auxiliary FER task would be quite limited.
To mitigate this issue, our proposed MAL increases the average weights of the FE samples as shown in Fig.~\ref{fig:bp4d_weight_loss} (e),(f),(i). It is evident that the meta net in MAL automatically balances the weights of the two tasks and  adaptively increases the average weights of the FE samples according to the potential of the iteration-weight curve.
It is reasonable because MAL learns to enhance the contribution of FER to extract more semantic information to boost the AU detection task. 

We additionally group the FE samples according to their facial expression categories and ranked the samples according to their weights. 
Fig.~\ref{fig:img_weight_vis} shows the results. It is clear that MAL adaptatively estimates low weights for ambiguous FE samples, and high weights for FE samples that have related AUs activated, e.g., AU4 (brow lower), AU5 (upper lid raiser) in angry,  AU9 (nose wrinkler) in disgust, AU1 (inner brow raiser) in fearful,  AU6 (cheek raiser), AU12 (lip corner puller) in happy, AU4 (brow lowerer), AU6 (cheek raiser) in sad, AU5 (upper lid raiser), AU26 (jaw drop) in surprised facial images. The FE samples with low weights usually have no AU activated, this is in line with the observation in \cite{yan2020raf}  that the  perceived emotion is distinguished by gaze, head pose, gesture, and facial appearance.  
For instance, the facial expressions in the second, fourth, fifth images in the top row in Fig.~\ref{fig:image_weight_consistency} (a) are categorized by their gazes, the facial expressions in the third images in the top row in (b), the first image in the top row in (d) are mainly distinguished by her gesture. 
The meta net in MAL is used as a gate to filter potentially less related FE samples which corresponds to the idea of importance reweighting \cite{cortes2010learning, shu2019meta}. The visualization results in Fig. \ref{fig:img_weight_vis} show our proposed MAL is capable of decreasing the contribution of the noisy samples in the auxiliary FER task and facilitating transferring more knowledge to boost the AU detection model.

Considering that the facial image in the RAF-DB dataset was independently annotated by 40 annotators,  we are capable of measuring the uncertainty of each image by calculating the proportion of the final labeled facial expression category in the overall 40 annotation results. We use the proportion as the consistency value for each FE sample. The larger the consistency value, the lower the uncertainty. We illustrate the weight and consistency value of some representative images in Fig.~\ref{fig:image_weight_consistency}. For each facial image in Fig.~\ref{fig:image_weight_consistency}, the two values on the top left mean the weight and consistency value. For each sub-figure in Fig.~\ref{fig:image_weight_consistency}, the top row means the images with small weight and the bottom row denotes the images with high weight. It is clear that the images with low consistency values usually have small weights, and vice versa. 
It indicates our proposed MAL suppresses the FE samples that have large uncertainties and prevents the network from over-fitting uncertain FE images. The learned sample weights are reasonable.

\subsection{Conclusion}
With this work we have presented a Meta Auxiliary Learning method (MAL) for facial AU detection. The proposed MAL adaptatively transfer knowledge from a large amount of FE samples to benefit AU detection by learning adaptative sample weights to filter the less related FE samples and balance the two tasks.  
Extensive experimental results show that MAL consistently improves the AU detection performance and is comparable with other state-of-the-art AU detection methods. 
Visualization experimental results show our proposed MAL is capable of estimating reasonable weights that not only balance the two tasks and automatically weigh the auxiliary samples according to their semantic relevance with the primary AU detection task.
For future work, we plan to utilize MAL in other multi-task learning scenarios.

% use section* for acknowledgment
% \ifCLASSOPTIONcompsoc
  % The Computer Society usually uses the plural form
%   \section*{Acknowledgments}
% \else
  % regular IEEE prefers the singular form
%   \section*{Acknowledgment}
% \fi

\ifCLASSOPTIONcaptionsoff
  \newpage
\fi

\bibliographystyle{IEEEtran} 
\bibliography{ieee-abrv}

\begin{IEEEbiography}[{\includegraphics[width=1in,height=1.25in,clip,keepaspectratio]{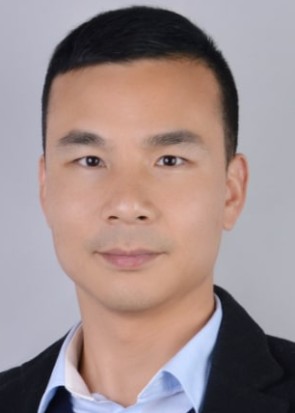}}]{Yong Li} has been an assistant professor at School of Computer Science and Engineering, Nanjing University of Science and Technology since 2020. His research interests include deep learning, self-supervised learning and affective computing.
\end{IEEEbiography}

\begin{IEEEbiography}[{\includegraphics[width=1in,height=1.25in,clip,keepaspectratio]{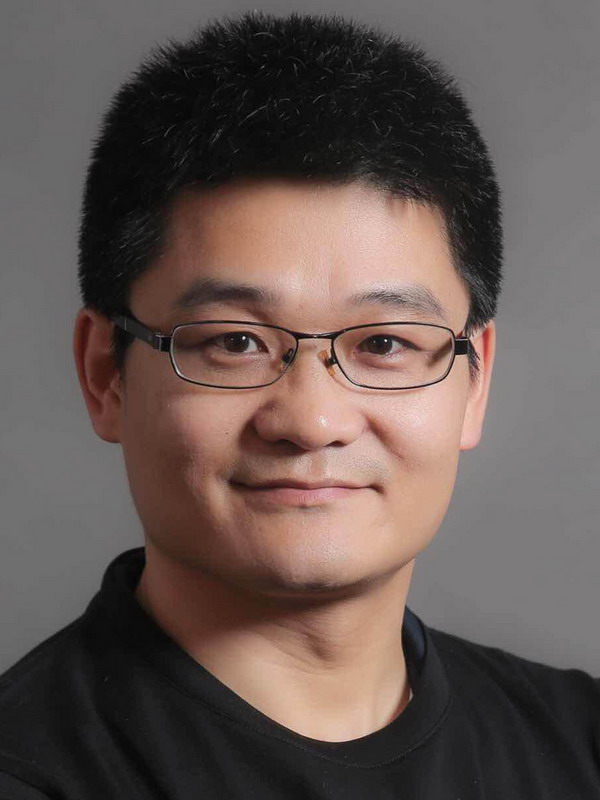}}]{Shiguang Shan} received M.S. degree in computer science from the Harbin Institute of Technology, Harbin, China, in 1999, and Ph.D. degree in computer science from the Institute of Computing Technology (ICT), Chinese Academy of Sciences (CAS), Beijing, China, in 2004. He joined ICT, CAS in 2002 and has been a Professor since 2010. He is now the deputy director of the Key Lab of Intelligent Information Processing of CAS. His research interests cover computer vision, pattern recognition, and machine learning. He especially focuses on face recognition related research topics. He has published more than 200 papers in refereed journals and proceedings in the areas of computer vision and pattern recognition. He has served as Area Chair for many international conferences including ICCV'11, ICPR'12, ACCV'12, FG'13, ICPR'14, ICASSP'14, ACCV'16, ACCV18, FG'18, and BTAS'18. He is Associate Editors of several international journals including IEEE Trans. on Image Processing, Computer Vision and Image Understanding, Neurocomputing, and Pattern Recognition Letters. He is a recipient of the China's State Natural Science Award in 2015, and the China’s State S\&T Progress Award in 2005 for his research work. He is also personally interested in brain science, cognitive neuroscience, as well as their interdisciplinary researche topics with AI.
\end{IEEEbiography}

% that's all folks
\end{document}